\documentclass[sigconf]{acmart}

\AtBeginDocument{%
  \providecommand\BibTeX{{%
    Bib\TeX}}}

\copyrightyear{2026}
\acmYear{2026}
\setcopyright{cc}
\setcctype{by-nc-nd}
\acmConference[FDG '26]{Foundations of Digital Games}{August 10--13, 2026}{Copenhagen, Denmark}
\acmBooktitle{Foundations of Digital Games (FDG '26), August 10--13, 2026, Copenhagen, Denmark}
\acmDOI{10.1145/3815598.3815605}
\acmISBN{979-8-4007-2495-4/2026/08}

\usepackage{amsmath,amsfonts}
\usepackage{algorithmic}
\usepackage{graphicx}
\usepackage{textcomp}

\usepackage{hyperref}

\def\BibTeX{{\rm B\kern-.05em{\sc i\kern-.025em b}\kern-.08em
    T\kern-.1667em\lower.7ex\hbox{E}\kern-.125emX}}

\usepackage{multirow}
\usepackage{makecell} 

\usepackage{colortbl}

\definecolor{golden}{HTML}{FFF8E1}   
\definecolor{fine16}{HTML}{E3F2FD}   
\definecolor{quant8}{HTML}{FFCDD2}   
\definecolor{quant4}{HTML}{C8E6C9}   

\usepackage{fancyvrb}
\usepackage{color}

\usepackage{stfloats}

\definecolor{boxborder}{gray}{0.55}
\definecolor{titlebar}{gray}{0.45} 

\begin{SaveVerbatim}[commandchars=\\\{\}]{inputdataverb}
\textbf{Sender}:
    \{'name': 'John Kantakouzenos', 'profession': 'knight',
    'personality_trait': ['Trustworthy', 'Objective', 'Pessimistic'],
    'faction': 'Order of the Golden Chalice',
    'description': 'John Kantakouzenos is a seasoned knight known for his unwavering loyalty to the Order of the Golden
    Chalice. His reputation for trustworthiness and objectivity precedes him, making him a respected figure among his
    peers. However, his tendency towards pessimism often casts a shadow over his otherwise noble demeanor.',
    'catchphrases': \{'public': 'For honor and the Golden Chalice!', 'private': 'Hope is but a fleeting shadow.'\}\}

\textbf{Target}:
    \{'name': 'Freydis Eiriksdottir', 'profession': 'nobility',
    'personality_trait': ['Open-minded', 'Competitive', 'Obnoxious'],
    'faction': "The Raven's Council",
    'description': "Freydis Eiriksdottir is a noblewoman of the Raven's Council, known for her adventurous spirit and
    strategic mind. She is both admired and feared for her competitive nature and her tendency to speak her mind, often
    to the discomfort of others.",
    'catchphrases': \{'public': 'The winds of change favor the bold.', 'private': 'If only these fools could see beyond
    their noses.'\}\}

\textbf{Intelligence}:
    \{'\textbf{failure}': "Freydis Eiriksdottir once attempted to lead a diplomatic mission to secure an alliance with a 
    neighboring faction, but her competitive nature led to a heated argument that resulted in the mission's failure.",
    '\textbf{addiction}': 'Freydis has developed an addiction to the thrill of high-stakes political maneuvering, often seeking
    out situations that allow her to exercise her strategic mind, regardless of the potential consequences.'\}
\textbf{Target Audience}: soldiers
\textbf{Angle}: bad family lineage
\end{SaveVerbatim} 

\begin{document}

\title{High-quality generation of dynamic game content via small language models: A proof of concept}

\author{Morten I. K. Munk}
\email{moim@itu.dk}
\affiliation{%
  \institution{brAIn lab\\IT University of Copenhagen}
  \city{Copenhagen}
  \country{Denmark}
}
\affiliation{%
  \institution{\&\\Raw Power Labs}
  \city{Copenhagen}
  \country{Denmark}
}

\author{Arturo Valdivia}
\email{arva@itu.dk}
\affiliation{%
  \institution{Data Science Section\\IT University of Copenhagen}
  \city{Copenhagen}
  \country{Denmark}
}

\author{Paolo Burelli}
\email{pabu@itu.dk}
\affiliation{%
  \institution{brAIn lab\\IT University of Copenhagen}
  \city{Copenhagen}
  \country{Denmark}
}

\begin{abstract}
Large language models (LLMs) offer promise for dynamic game content generation, but they face critical barriers, including narrative incoherence and high operational costs. Due to their large size, they are often accessed in the cloud, limiting their application in offline games. Many of these practical issues are solved by pivoting to small language models (SLMs), but existing studies using SLMs have resulted in poor output quality. We propose a strategy of achieving high-quality SLM generation through aggressive fine-tuning on deliberately scoped tasks with narrow context, constrained structure, or both. In short, more difficult tasks require narrower scope and higher specialization to the training corpus.  Training data is synthetically generated via a DAG-based approach, grounding models in the specific game world. Such models can form the basis for agentic networks designed around the narratological framework at hand, representing a more practical and robust solution than cloud-dependent LLMs.
To validate this approach, we present a proof-of-concept focusing on a single specialized SLM as the fundamental building block. We introduce a minimal RPG loop revolving around rhetorical battles of reputations, powered by this model. We demonstrate that a simple retry-until-success strategy reaches adequate quality (as defined by a rubric-based LLM-as-a-judge scheme) with predictable latency suitable for real-time generation. Generation time estimates based on human annotation and cross-model validation suggest that the retry strategy remains practical even under substantially stricter quality requirements for the quantized models. While local quality assessment remains an open question, our results demonstrate feasibility for real-time generation under typical game engine constraints.
\end{abstract}  

\maketitle

\section{Introduction}

Artificial intelligence has long been integral to the video game industry~\cite{Yannakakis2025}, with a particular emphasis on adaptive narrative elements such as believable NPCs~\cite{Cass2002MindGames,Merrick2007ModelingMotivation}. Since game elements that provide clear feedback and consistently respond to player agency correlate with increased engagement~\cite{Klimmt2007EffectanceControl,SchoenauFog2011PlayerEngagement}, recent advances in generative AI have naturally opened up new avenues for dynamic content generation using large language models (LLMs), including explorations of NPC dialogue~\cite{parkGenerativeAgents2023,plougNPCDialogue2025} and quest generation~\cite{AlNassar2023QuestVille}. In this paper, we explore an alternative paradigm for narrative AI in games, which we believe is more suited for the practical reality of integration into video games: aggressively fine-tuned small language models (SLMs), deliberately constrained to narrow, well-defined generation tasks, potentially arranged in agentic networks.

The paper is organized as follows: Section~\ref{sec:ProposedFramework} outlines our proposed framework and the technical constraints under consideration. Section~\ref{sec:DefameLM} presents DefameLM, our proof-of-concept implementation, and details both the game loop design and model training. Section~\ref{sec:Results} contains a quantitative evaluation of DefameLM's output quality at different levels of quantization, as well as the generation efficiency. In Section~\ref{sec:Discussion} we discuss limitations and future directions, and in Section~\ref{sec:Conclusion} offer conclusions.

\subsection{Obstacles of LLMs for game content}\label{sec:HistoryLLMinGames}
Although holding much promise, LLMs appear to have difficulty forming a coherent understanding of complicated worlds, as exemplified by Tsai \emph{et al.}, who showed that a straightforward application of then-state-of-the-art LLMs is insufficient for playing a text-based game, \emph{Zork I} (1977)~\cite{Tsai2023CanLargeLanguage}. Specifically, they tasked ChatGPT-4 with playing the game with access to the game manual as well as few-shot examples, but cited an inability of the LLM to successfully infer and utilize knowledge about the game world and formulate goals. This suggests that naive applications of LLMs to control NPCs in more complicated game situations will likely fail to adhere to the world and maintain logical coherence. Since maintaining narrative comprehension is essential for player immersion in story-driven games~\cite{qinMeasuringPlayerImmersion2009}, the LLMs' difficulty in maintaining narrative cohesion underscores a significant challenge to LLM integration in narrative-driven games. Müller-Brockhausen \emph{et al.} found that even superficial "chatter" (or "barks" in game developer terminology) can struggle to stay in context, highlighting the challenge of full dynamic NPC dialogue~\cite{muller-brockhausenChatter2023}.

Indeed, lack of consistency and guard-railing of LLMs is an often cited shortcoming for AI generated narrative content~\cite{Gallotta2024,Sweetser2024}. Maintaining narrative consistency requires LLMs to track extensive game state and character information, which can be managed through techniques such as RAG or complex prompting. However, increasing prompt complexity and constraint specificity beyond a certain point has been shown to reduce narrative coherence and creativity~\cite{Atmakuru2024}. This creates tension between the need for consistency (requiring more context/constraints) and the need for creative, engaging narratives.

\subsection{Are agentic LLM frameworks the solution?}\label{sec:AgenticFrameworks}
Recent studies have suggested that some of the above challenges can be mitigated by adopting an \emph{agentic} framework, where complex tasks decompose into smaller subtasks handled by separate LLM calls, reminiscent of AI chaining~\cite{Wu2022AIChains} and chain-of-thought reasoning~\cite{Zhang2022AutomaticChainThought,Wei2022Chainofthought}. For example, a complex NPC decision can be decomposed into subtasks such as summarizing recent events and evaluating the NPC's psychological state, some of which can run ahead of time. Jeong \emph{et al.}~\cite{Jeong2025} applied this approach and found high player engagement but persistent coherence problems and a lack of direction when no explicit goals were provided to the LLM. This points towards a fundamental challenge: narratives comprise incompatible operations, and generating across them requires careful analysis of the specific use-case. Gervás \emph{et al.}~\cite{Gervas2019} stressed exactly this, proposing to separate concerns such as plot structure from character behaviour into distinct LLM calls. Conversely, handling tasks monolithically would require either encoding the entire narrative framework into the prompt, or relying entirely on the LLM to make such decisions. For instance, the logic for tracking and developing narrative tension, managing valid transitions, and triggering character development must be handled while maintaining consistency across all these dimensions. An example of this difficulty is quest generation, which touches all these aspects simultaneously. Encoding them into a single prompt would be a monumental task, and the alternative, relying on the LLM's own reasoning, cedes creative control and assumes it can maintain a continuous coherent understanding of the entire game state. Decomposition into separate calls offers a more dependable and controllable approach. A recent example of story generation using deliberate narrative structure is given by Wen \emph{et al.}~\cite{Wen2025AllStoriesAreOneStory}, who constructed coherent branching stories using a directed acyclic graph (DAG) approach combined with a reductionist story framework of rising and falling narrative beats, finding that recognizable narratological structure significantly increased player immersion. However, synchronizing generated narrative with gameplay remained an open challenge, illustrating that narrative generation designed independently of gameplay elements cannot be expected to correlate meaningfully with them. Where such correlation is intended, designing the two in tandem may be necessary.

Despite these advances and the increasing capabilities of state-of-the-art LLMs, they still see very limited application in commercial video games. Indeed, some barriers are fundamental: cloud-based LLMs make single-player games online-only, introduce unpredictable costs for either the studio or the player, and are vulnerable to server shutdowns, model deprecation, or replacement, rendering game features non-functional or unpredictable. This work specifically addresses local deployment scenarios, making comparison to cloud-based APIs out-of-scope. 

\section{Proposed framework: Agentic system of SLMs}\label{sec:ProposedFramework}

In light of the move toward narrowly scoped, agentic LLM frameworks in games, we propose addressing the remaining practical barriers by replacing monolithic LLM calls with an agentic network of task-specific fine-tuned SLMs. By 'agentic,' we refer to a DAG organization where each node handles a narrowly defined subtask, e.g. generation for a specific purpose, retrieval, or function calls such as altering game states or directing NPCs. The key principle is specialization: Each SLM is aggressively fine-tuned for a single, well-defined task rather than attempting to handle complex, multi-faceted operations. This approach aligns with emerging industry interest in SLMs for agentic systems. Outside of gaming, this paradigm has shown promise as a flexible, private, durable, sustainable, and economical approach for agentic AI applications, as demonstrated in recent NVIDIA research~\cite{Belcak2025SLMAgentic}.

Crucially, for generation of short, clearly contextualized creative content, SLMs have been shown to achieve comparable quality with LLMs and human writers~\cite{Marco2025SLMOutperformHumans}. While the use of fine-tuned SLMs holds significant promise, their application in the context of dynamic game content remains largely unexplored. 
Värtinen \emph{et al.} attempted monolithic quest generation via SLM fine-tuning on 978 samples, yielding inconsistent results~\cite{Vartinen2024GeneratingRPG}. We suspect that decomposition is necessary rather than merely beneficial for tasks of this complexity, even when using LLMs, and that agentic decomposition into specialized SLMs could substantially improve such results. In our SLM framework, robust guard-railing and adherence to world constraints, tone, and structure are achieved through deliberate training data curation. For difficult tasks, guidance is necessary to achieve quality, creating a \emph{creativity-consistency trade-off}, especially important when models become small enough or tasks become sufficiently complex. Our approach provides two control mechanisms: data variety and degree of fitting to the training data. Unlike LLMs where small prompt changes yield unpredictable results, SLM fine-tuning offers more predictable control since these parameters directly determine how closely the model replicates the training dataset's style and structure. However, this predictability requires fine-grained control over dataset generation.

Complex tasks such as dialogue can potentially touch on many disparate facets of gameplay, such as the general world state, the current goals of the NPC, the player's active and completed quests, and the current activity the player is engaged in, but dynamic content generation does not always require handling this full complexity. We define task complexity by first identifying its constituent \emph{base dimensions}: independently definable aspects of the task such as narratological requirements (plot coherence, character consistency and development, etc.), stylistic constraints (tone, genre adherence, humor), and cognitive demands (logical synthesis, information sorting, contextual reasoning). The base dimensions may be highly correlated, for example, if an output must be humorous based on a character trait as well as a plot development. We define complexity as the number of correlated base-dimension tuples that the task requires to be satisfied jointly, and refer to each such tuple as a \emph{complexity dimension}. Given that the pervasive challenge with dynamically generated content is coherence, much of the
complexity arises from  correlations between base dimensions that shift depending on the context of the generation. Distinguishing the context itself can be a high-complexity task, compounding the
difficulty. In practice, therefore, we can distinguish between two types of generation contexts, each potentially calling for a different approach: 
\begin{enumerate}
    \item \textbf{Open-ended contexts}, such as generic NPC dialogue, where a complex task may have to be decomposed into smaller subtasks, each handled by a specialized model. 
    \item \textbf{Game-loop-anchored contexts}, where the generation is confined to an explicitly defined game situation or loop and a single more narrowly scoped model may be sufficient.
\end{enumerate}

Both approaches align with our proposed framework of specialized SLMs composed into agentic networks. The former may require multiple coordinated models but potentially enables more general-purpose generation. The latter is automatically contextualized and more readily measurable, and thus the framework can be stress-tested by focusing on such tasks. In particular, by targeting scenarios where the generative task is challenging (requiring wit, synthesis, contextual awareness) but the scope remains narrow enough for aggressive fine-tuning to reliably produce quality outputs, we can investigate if the most fundamentally difficult tasks are feasible in this framework. 
If such challenging tasks can furthermore be solved by a single fine-tuned SLM, this represents the most fundamental demonstration of our framework. Establishing this viability also offers immediate practical value by giving designers a methodology for incorporating dynamic content: either by identifying suitable existing game loops or designing them from scratch, and then using an SLM to interpret and contextualize the related events within the game world's narrative framework. Anchoring generation within a loop additionally addresses the synchronization challenge observed by Wen \emph{et al.} A loop-centric approach, when applicable, ensures that generated content and gameplay logic remain tightly coupled. Game-loop-anchored contexts can still entail high complexity when requiring synthesis of large information sets, inference of intentions, and simultaneous satisfaction of multiple constraints while maintaining appropriate tone and humor. For the remainder of this paper, we turn our attention to demonstrating this proof of concept.

\subsection{Minimal proof of concept}\label{sec:ProofOfConcept}
We consider an RPG interaction loop centered around a \emph{reputational conflict} between two characters. We present \emph{DefameLM}, the single fine-tuned SLM servicing the entire game loop by generating textual outputs of rhetorical blows between the characters, specifically in the form of propaganda fitting for posters in a medieval market place. This game loop has a fixed level of complexity: the task always requires synthesizing 1-2 intelligence items, implementing a rhetorical angle, targeting a specific audience, and finding a humorous take all the while maintaining medieval setting consistency within approximately 150 words. By showing that one self-contained SLM can service an entire gameplay loop, we highlight the approach's immediate viability.

To establish infrastructure-level feasibility, we evaluate whether DefameLM can successfully execute the complex task (synthesizing intelligence, implementing rhetorical angles, targeting audiences, maintaining tone and consistency) to a quality standard comparable to the training data. We assess each task requirement individually, explicitly checking overlapping dimensions, and failure on any single test constitutes overall failure. The models are evaluated at three quantization levels (16-, 8-, and 4-bit), measuring both generation quality and time-to-success under a retry-until-success strategy, where success is defined by the verdict of a suite of LLM-as-judge calls. When generation failures occur, our evidence suggests they are predominantly recoverable rather than deterministic. By introducing stochasticity ($T=0.75$), in a small number of retries the model is able  to explore variations until an acceptable output is found. As demonstrated in Section~\ref{sec:Results}, this approach yields predictable success rates and latency suitable for real-time generation in the majority of cases, even for the most heavily quantized version. 

\subsection{Technical constraints for real-time generation}\label{sec:TechnicalConstraints}
We restrict our consideration to applications where content generation can be masked by pre-scripted sequences such as brief dialogue or cutscenes, establishing practical constraints for evaluating DefameLM. For such use cases, we define a generation to be viable if it completes within a time no more than approximately $5$ seconds. Within this window, memory constraints become manageable since models can be loaded during these frozen states and unloaded afterward. The exact memory requirements vary significantly by implementation and represent a moving target throughout game development. Consumer GPUs typically offer around 8GB of VRAM, but resource-intensive games with demanding graphics and complex engines leave limited headroom. Models exceeding 2-3 GB therefore become increasingly difficult to accommodate, though the acceptable threshold depends heavily on the specific game's architecture and optimization priorities. For evaluating our proof-of-concept, we target sub-2GB models as a practical benchmark, while acknowledging that even smaller footprints may be necessary for particularly resource-constrained scenarios. These conditions establish the performance constraint within which we evaluate our implementation in Section~\ref{sec:Results}.

\section{DefameLM}\label{sec:DefameLM}
We now turn to the implementation of DefameLM. In this section we will introduce its game loop, the training data generation process, and the model training itself. 

\begin{figure}
    \centering
    \includegraphics[width=1\linewidth] {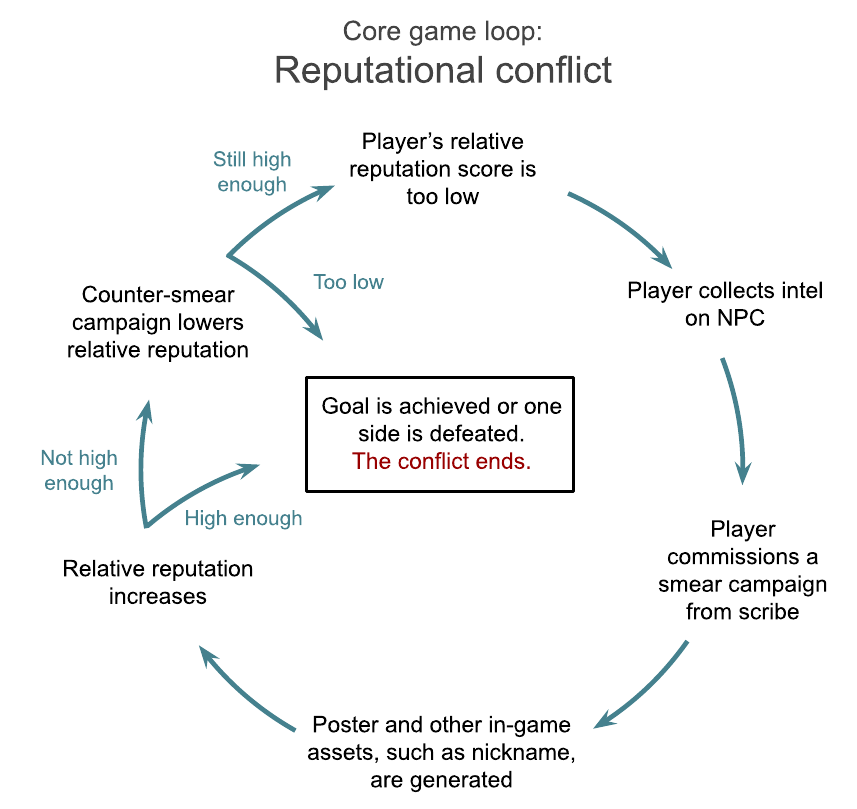} 
    \caption{Game loop centered on reputational conflict where participants pursue goals (e.g., inciting a riot) blocked by insufficient relative reputation. Players collect intel, initiate smear campaigns, and manage reputation to outmaneuver opponents, who can retaliate with their own smear campaigns. The loop continues until one side achieves its goal or is defeated, ending the conflict.}
    \label{fig:GameLoop}
\end{figure}

\begin{figure*}[!t]
    \centering    \includegraphics[width=.775\linewidth]{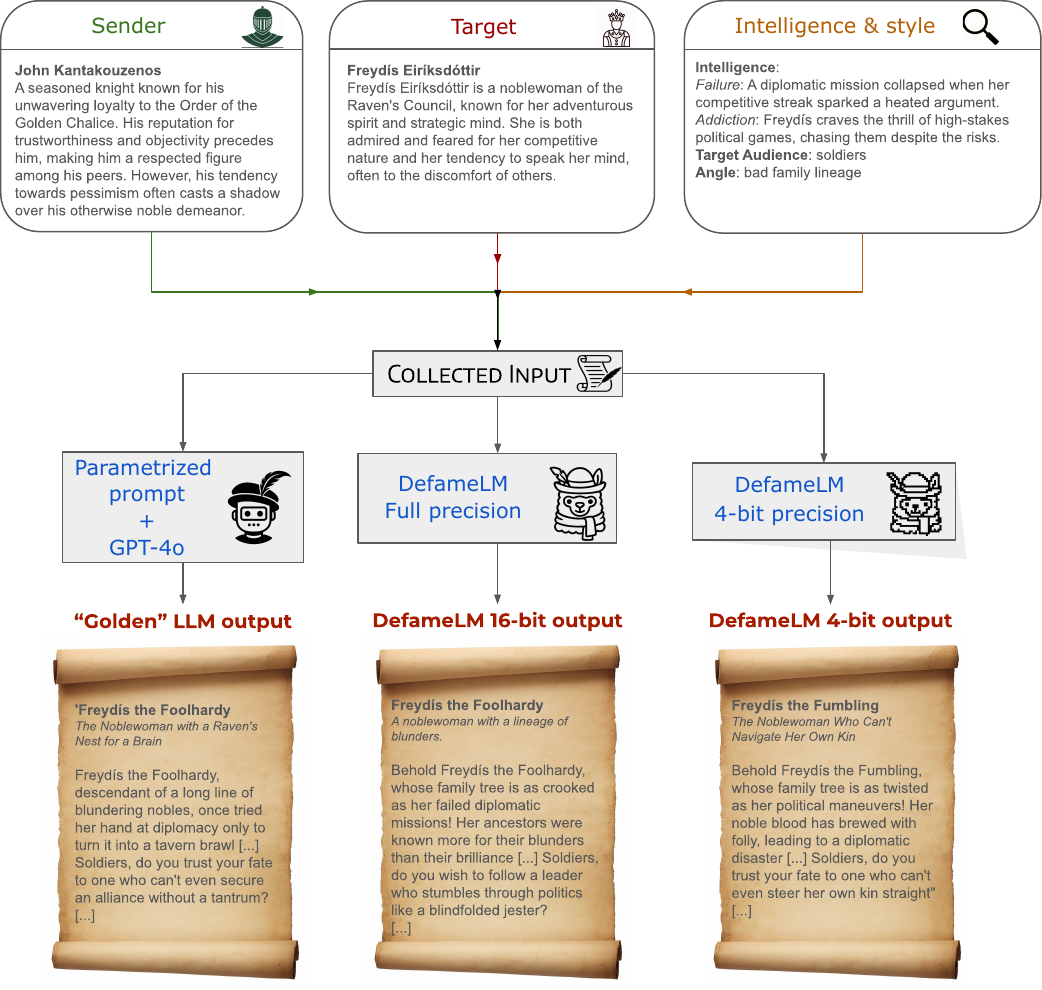}
    \caption{Input/output structure for DefameLM with example outputs from GPT-4o (training data gold standard), DefameLM 16-bit, and DefameLM 4-bit. The model takes as input basic information about the sender and target (either NPC or player), plus contextual information: 1) intelligence about the target, 2) the target audience, and 3) the rhetorical angle of the smear campaign. Inputs shown are shortened, full outputs appear in Appendix \ref{app:FullTextSample}. DefameLM was trained without instruction prompting to associate output structure with raw JSON inputs containing sender, target, intelligence, and style data.}    \label{fig:InputOutputDefameLM}
\end{figure*}

\subsection{Game core loop: The reputational conflict}\label{sec:GameLoop}
 
The game loop illustrated in Fig.~\ref{fig:GameLoop} is designed for a hypothetical RPG focused on building power and reputation, either militarily by raising an army, or maybe through influence, espionage and diplomacy. Reputation may be earned organically through player actions or deliberately shaped by crafting written content, such as books, letters, or smear campaigns. In this paper, we deal with the latter option. Thus, the game loop consists of the player gathering information, rumors, lies and the like, maybe through infiltration, bribery, burglary, blackmail or simply by being at the right place at the right time. This information is then taken to a scribe of dubious moral quality, who writes a poster to be hung around town. Mechanically, one or more numerical reputation scores are updated deterministically (without AI), reflecting factors such as the effectiveness of the player's strategic choices for the intended audience. This can affect the game world in various ways, such as whether the player can hire mercenaries from under the nose of a previously popular warlord, or whether war is triggered or avoided.

The fine-tuned SLM generates these scribe outputs. The input/output structure is shown in Fig.~\ref{fig:InputOutputDefameLM}. A \emph{sender} writes a smear campaign about a \emph{target}, where either may be the player or an NPC. The attack is based on unlockable \emph{intelligence} and stylistic elements. The text becomes in-game assets: posters and extractable elements like invented nicknames that NPCs reference dynamically.

The model receives character metadata (e.g., faction allegiance, physical appearance, general backstory, and personality) along with the intelligence, which consists of three parts: one or two compromising pieces of information, the target audience of the smear, and a rhetorical angle, that is, the comedic lens through which the target is attacked, such as bad fashion sense or illiteracy. The model is trained to aggrandize the sender and belittle the target, with rhetorical choices tuned to appeal to the selected audience. The intelligence may include gameplay-derived details --- recent defeats, cancelled construction projects, behavioral patterns like talking to every NPC --- translated into natural language inputs. By appearing in propaganda, such actions receive emergent narrative contextualization: frequent cancellations of building projects for a player in the mason's guild might become "hypocritically never paying contractors", or a completionist playstyle might be characterized as "sticking their nose in everyone's business". This expands what games can narratively acknowledge since arbitrary player actions can be dynamically incorporated into the game content.

DefameLM is trained on samples generated by GPT-4o, and we study it at three levels of quantization (16-, 8-, and 4-bit). The parametrized prompt constrains the outputs to have a length of no more than 500 characters, and the fine-tuning of DefameLM effectively captures this constraint for all quantization levels. The length limit makes it a fairly significant challenge to give due weight to all the parts of the input. Figure 2 contains a sample input (edited for brevity) alongside the corresponding outputs from GPT-4o and all three quantization levels of DefameLM, for the full texts see App.~\ref{app:FullTextSample}. Although the quality of the 16-bit version is usually high, the 4-bit model occasionally exhibits a sizeable drop in quality, usually by making an illogical connection, forgetting tasks or making failed punchlines or metaphors.

\subsection{Generating the training data: DAG approach}\label{sec:DataGeneration}
To achieve fine-grained control over dataset generation, we need sufficient variability for the model to generalize while maintaining focus on game-specific elements for quality generation. We designed a DAG-based iterative process that provides this control. In essence, we use a larger teacher LLM to generate text samples for fine-tuning, where each piece follows a structured multi-step generation scheme. Once the input format and parametrized prompt $\Pi$ are fixed, our DAG approach enables systematic control over the variety of training input/output pairs while ensuring adherence to game lore.

\begin{figure}[t!]
    \centering
    \includegraphics[width=1\linewidth]{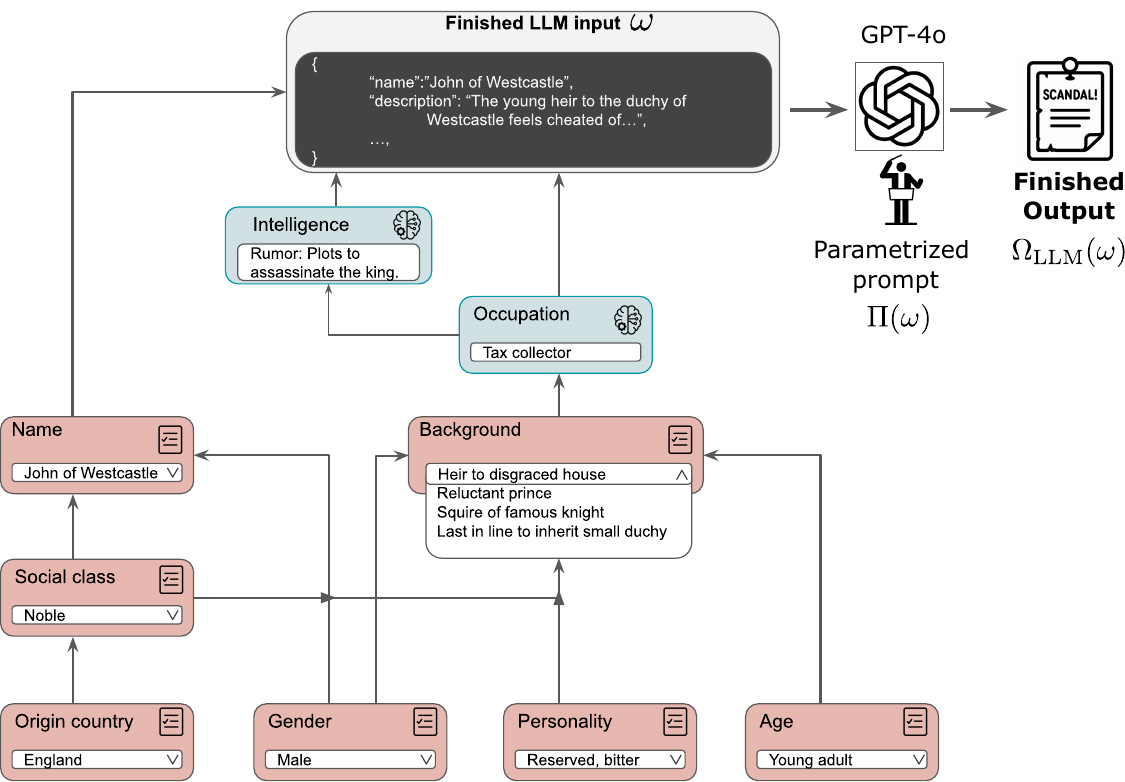}
    \caption{Example of the DAG-based data generation using choice nodes (predefined lists) and generation nodes (LLM-generated content):  A finished LLM input is stitched together from various parts that are either chosen from conditional lists, dependent on earlier choices, or generated by an LLM. In our example, all conditional lists were generated ahead of time using GPT-4o, and generations during the DAG execution were carried out using GPT-4o-mini. The final collected input $\omega$ populates a parametrized prompt $\Pi(\omega)$, which is sent to a final GPT-4o call, thus producing the final LLM output $\Omega_{\text{LLM}}(\omega)$.}
    \label{fig:DataGenerationDAG}
\end{figure}

We first decompose the inputs $\omega$ into a list $\vec{z}$ containing details such as the origin country, personality, factional allegiance, and occupation of the sender and target. This information is assumed to be read from the game when later generating during runtime. To avoid redundancy in the collection of inputs, we employ an iterative process, whose logic follows a DAG. The process is exemplified in Fig.~\ref{fig:DataGenerationDAG}. Each component of $\vec{z}$ is either selected from an appropriate predefined list (\emph{choice nodes}) or generated using any elements up until the given point in the DAG execution (\emph{generation nodes}). The lists may be nested, such that origin country may affect which social classes a character can belong to, for example. Choices at any given point in the DAG flow may affect which lists are available downstream. Variation is guaranteed by selecting different combinations of all the lists and stitching together the final input $\omega$ using GPT-4o-mini to write description and other free-form texts specific to all the collections of choices and generations. The approach is a powerful way of achieving a large quantity of text samples grounded in the game lore at hand. 

The final input $\omega$ is used to populate the parametrized prompt $\Pi$. The resulting prompt $\Pi(\omega)$ is sent to the teacher model, GPT-4o~\cite{openai2024GPT4,openai2024GPT4o}, generating the final training output $\Omega_{\text{LLM}}\big(\Pi(\omega)\big)$. This pipeline, akin to self-instruction~\cite{Wang2023SelfInstruct}, was used to generate 1800 distinct input-output pairs.

\subsection{Training DefameLM model}\label{sec:Training}
The model is obtained through task-specific supervised fine-tuning of a small base model, Llama 3.2-1B~\cite{Grattafiori2024Llama3}, using the synthetically generated data. Specifically, we employ low-rank adaptation (LoRA)~\cite{Hu2021LoRA} of the base model, using a prompt-loss weight~\cite{HuertaEnochian2024PromptLossWeight} of $5\,\%$.

\begin{figure}[t]
    \centering
    \includegraphics[width=1\linewidth]{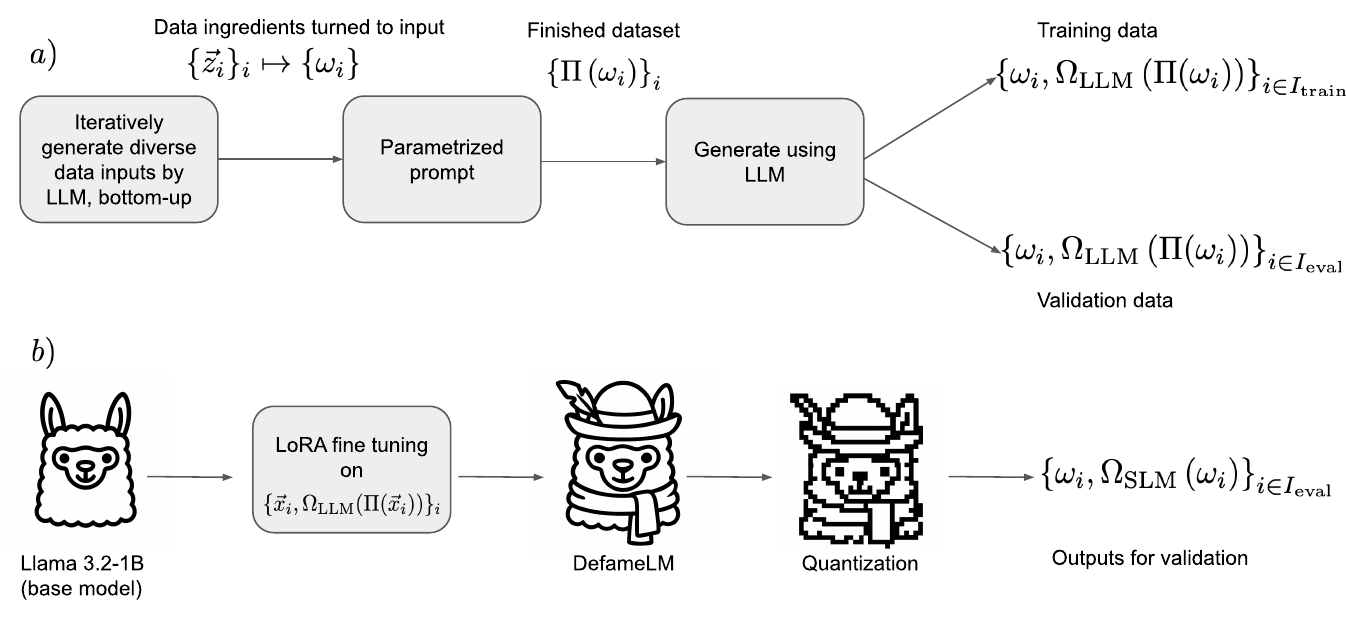}
    \caption{Training data generation and fine-tuning pipeline. $a)$ Varied inputs $\{\vec{z}_i\}_i$ are generated following the structure of Fig.~\ref{fig:DataGenerationDAG}, mapped to string inputs $\omega_i$ populating the parametrized prompt $\Pi$. LLM generation produces outputs $\Omega_{\text{LLM}}(\Pi(\omega))$. $80\,\%$ of the inputs form the training dataset $I_{\text{train}}$, the remainder forms the validation set $I_{\text{eval}}$. $b)$ Base model Llama 3.2-1B is fine-tuned on training data, yielding 16-bit (training precision), 8-bit, and 4-bit quantized versions generating outputs $\Omega_{\text{SLM}}(\omega_i)$.}
    \label{fig:Model_training}
\end{figure}

From the 1800 generated pairs, we randomly selected 1440 inputs for fine-tuning and reserved the remaining 360 for evaluation. Let $I_{\text{train}}$ and $I_{\text{eval}}$ denote the training and evaluation index sets respectively. The training process is outlined in Fig.~\ref{fig:Model_training}. The base Llama model is trained on the pairs $\{\omega_i, \Omega_{\text{LLM}}(\Pi(\omega_i))\}_{i \in I_{\text{train}}}$. As such, during generation the resulting fine-tuned model takes only $\omega$ as input, while the structure of the original prompt template has been learned during training.

The fine-tuning was performed using the Hugging Face SFTTrainer library~\cite{vonwerra2022trl}, with a learning rate of $2\cdot10^{-5}$, with a LoRA $\alpha=128$ and rank $r=256$. Since size and speed are paramount for our application, the resulting model is produced and compared at three different levels of quantization: 4-, 8- and 16 bit. With the base model in question, the corresponding memory footprints are: 16-bit (2.48 GB), 8-bit (1.32 GB), and 4-bit (808 MB). The 16-bit model exceeds our 2GB guideline from Section~\ref{sec:TechnicalConstraints}, while the 8-bit and particularly the 4-bit models fall within the target range for practical deployment alongside demanding game systems.

\section{Quantitative assessment of quality}\label{sec:Results}
Having established the game loop and a gold standard through the teacher LLM, we now quantitatively assess the quality of the SLM outputs relative to this standard.
The quality of DefameLM outputs varies considerably across input prompts, especially for the smaller models. Common failure modes include failing to capture
all provided intelligence, senseless metaphors, poorly implementing the rhetorical angle,
or making illogical connections when attempting to synthesize
disparate details. When a non-zero temperature is applied, quality fluctuates across generations, and often a successful output can be generated even when the zero-temperature output failed. We would like to diagnose the root cause of failures, investigating whether they are always recoverable through stochastic variation or whether failing outputs correspond to inputs in "dead zones" of the input space, where the model does not know what to do and struggles consistently regardless of temperature. Ultimately we would like to investigate whether the necessary output quality can \emph{in principle} be reached in reasonable time, as defined in Section~\ref{sec:TechnicalConstraints}, across prompts.

\subsection{LLM-as-a-judge experiment}
To assess this, we utilize a type of LLM-as-a-judge procedure~\cite{Zheng2023JudgingLLMasaJudge,Li2025Judge}, specifically a rubric-based technique evaluating outputs against a suite $S_J$ criteria, one for each complexity dimension. Since the SLM cannot be expected to exceed its teacher's quality, we use GPT-4o as both teacher and judge, calibrating the quality bar to the standard the SLM is trying to replicate. A concern with this approach is that self-bias may make GPT-4o irrationally lenient towards the SLM outputs. But human annotation and cross-model validation indicate that the self-bias only weakly extends to the student models. We refer to Section~\ref{sec:Limitations} for details. Each criterion $m \in S_J$ is scored as pass/fail ($j_m(\omega) = 0,1$, respectively), using a criterion-dependent judge prompt $\pi^m_J$. Hence, we rely on the LLM's absolute measure of quality, rather than relative rankings. Here and in the following, for brevity we suppress the judge score dependency on the LLM prompt, judge prompt and the SLM output, writing $j_m(\omega) := j_m\big(\pi^m_J,\Omega_{\text{SLM}}(\omega), \Pi(\omega)\big)$. 

The overall verdict of the LLM-as-a-judge is given by  
\begin{equation}\label{eq:judge_verdict}
J(\omega) = \min_{m \in S_J} j_m(\omega).
\end{equation}
In other words, an output is graded as pass only if it passes all tests. The tests used are

\begin{enumerate}
    \item \textbf{Overall assessment}: Does the output follow from the instructions in the prompt $\Pi(\omega)$? 
    \item \textbf{Angle implementation}: Does the text successfully convey the intended thematic or humorous angle?
    \item \textbf{Intelligence implementation}: Are all pieces of intelligence incorporated in a coherent and meaningful way?
    \item \textbf{Alignment}: Does the text adhere to the given constraints, remain within the medieval world, and avoid factual hallucinations or inappropriate language?
    \item \textbf{Writing quality}: Is the text well-written and stylistically consistent with a medieval RPG setting?
    \item \textbf{Audience targeting}: Does the text effectively tailor its content and tone to the intended audience group (e.g., peasants, nobles, guards)?
    \item \textbf{Rhetorical targeting}: Is the outputted smear campaign successfully directed at the intended target?
\end{enumerate}

\begin{figure}
    \centering
    \includegraphics[width=1\linewidth]{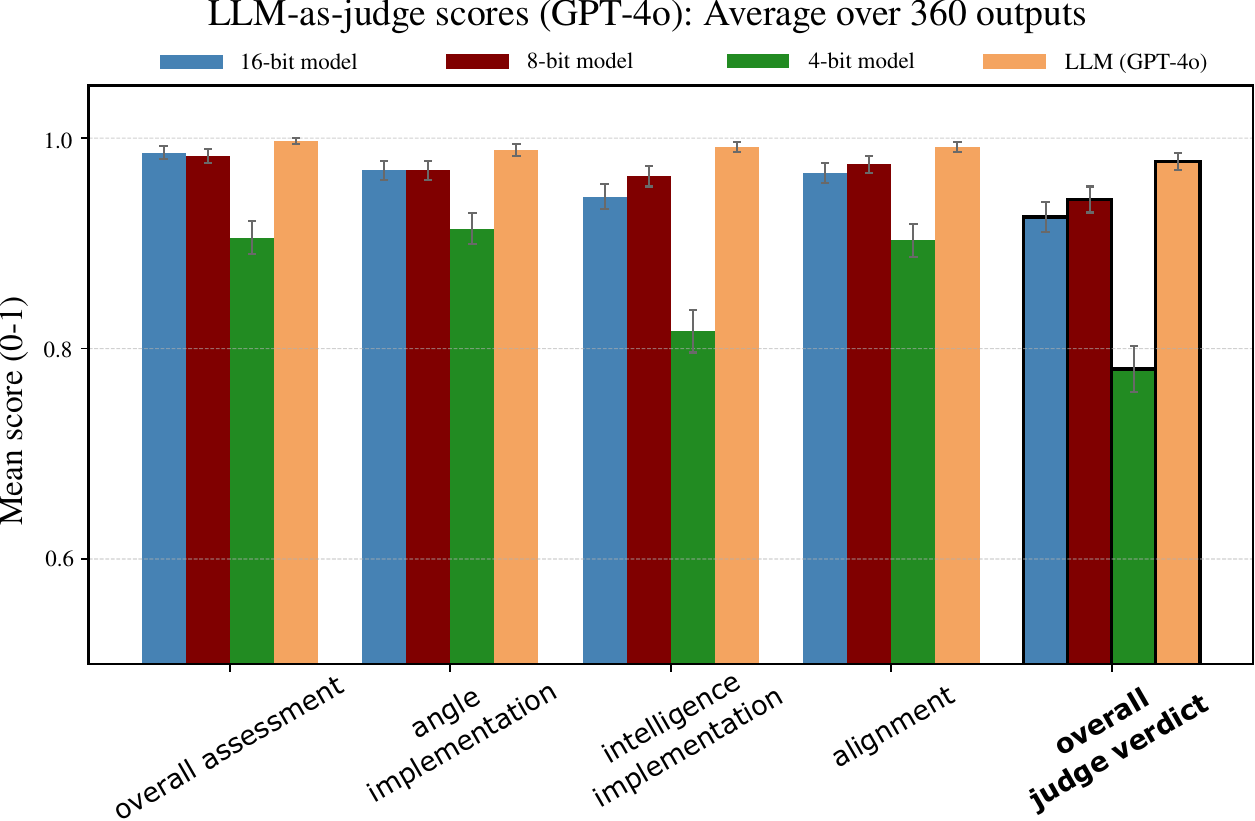}
    \caption{Mean LLM-as-judge scores per metric for DefameLM at different quantization levels (16-, 8-, and 4-bit) compared to the GPT-4o baseline (gold standard training data), averaged over 360 test samples. Three metrics (\emph{writing quality}, \emph{audience targeting}, and \emph{rhetorical targeting}) are omitted as all models achieved perfect scores. The \emph{overall judge verdict} (rightmost) represents the minimum across all metrics, requiring all individual tests to pass. Error bars show standard error ($\sigma/\sqrt{N}$). Despite the 16-bit model scoring 2 percentage points lower than the 8-bit model on overall verdict, overlapping confidence intervals show that this difference is not statistically significant.
    }
    \label{fig:AllJudgeResults}
\end{figure}

The overlap among criteria 1--4 reflects the correlation of the underlying complexity dimensions: a failure in intelligence implementation may simultaneously manifest as a weak rhetorical angle, and only overlapping criteria can catch such correlated failures. The overall verdict requires all tests to pass, so overlaps of criteria also provide a measure of protection against false positives. Criteria 5--7 achieved a perfect pass rate under GPT-4o across all models. To retain a measure of safeguarding on these dimensions while reducing cost in the per-prompt variation experiment (Section~\ref{sec:ECDF}), we used GPT-4o-mini for these criteria. Fig.~\ref{fig:AllJudgeResults} shows the overall performance of the different models under criteria 1--4, as well as the overall verdict, averaged over all 360 test samples. These outputs were generated with a temperature of $T=0.75$, using a top-$p$ strategy of $p=0.9$. The overall success rate of the LLM is a near-perfect  $98\,\%$. Using the standard error to estimate the accuracy of the success probabilities, the 8-bit model achieved $94.2\,\%\,\pm 1.2\,\%$ versus $92.5\,\% \pm 1.4\,\%$ 16-bit. While the 8-bit model's average score is slightly higher, the confidence intervals overlap, and McNemar's test yields $\chi^2 = 0.69$, $p = 0.41$. We conclude that the models are statistically indistinguishable, validating the 8-bit model as a drop-in replacement. 
The 4-bit model shows a clear degradation of quality, with a success probability of $78\% \, \pm 2.2\,\%$. Although the 4-bit model statistically needs more attempts to succeed, its faster inference speed may compensate. However, the number of expected attempts presumably fluctuates across prompts, and it is not \emph{a priori} apparent that the 4-bit model does not develop catastrophic failure modes on the hardest inputs, which would render the retry strategy impractical. We now turn to a quantitative investigation of these concerns.

\subsection{Estimating input-dependent success probability}\label{sec:ECDF}
We want to assess whether the models are always capable of generating appropriate output over a range of different inputs in an appropriate time on consumer hardware. To do this, we take $50$ random inputs, and for each of the $3$ levels of quantization we generate $100$ outputs using the same generation parameters as above. All outputs are scored by the LLM-as-a-judge rubric. For a given input $\omega_i$ and model $M$, we can estimate a success probability $p_i^M$. With this, consider the Bernoulli process of generating until success is achieved. The expected ''waiting time'' $W_i^M$, i.e. the average number of attempts before success, is~\cite{Feller1968}

\begin{equation}\label{Eq:WaitingTime}
    W_i^M = 1/p_i^M.
\end{equation}
We use a generation temperature of $T=0.75$, chosen as a deliberate compromise. When deterministic generation ($T=0$) produces an output that fails the judging criteria, introducing stochasticity allows the model to explore variations, some of which may succeed. However, a temperature that is too low provides insufficient variation to escape failures, while too high a temperature degrades overall quality. Preliminary inspection suggested that $T=0.75$ offered a reasonable balance between exploring creative variations and maintaining narrative coherence. A thorough optimization of this hyperparameter is beyond the scope of this study.

\begin{table}[h]
\centering
\caption{Median generation timing breakdown on consumer hardware. }
\label{tab:timing_breakdown}
\small
\begin{tabular}{lccc}
\hline
\textbf{Component} (median) & \textbf{4-bit} & \textbf{8-bit} & \textbf{16-bit} \\
\hline
Model loading ($\bar{t}_{\text{prep}}$) & 1145 ms & 1397 ms & 1782 ms \\
Single-attempt total ($\bar{t}_{\text{eval}}$) & 678 ms & 994 ms & 2784 ms \\
\hline
Per-token generation & 3.6 ms & 5.1 ms & 16 ms \\
\hline
\end{tabular}
\end{table}

\begin{table}[h]
\centering
\caption{Expected attempts and time-to-success percentiles across 50 test prompts (100 samples each).}
\label{tab:time_to_success_percentiles}
\small
\begin{tabular}{lcccccc}
\hline
\textbf{Model} & \textbf{Min} & \textbf{P25} & \textbf{Med} & \textbf{P75} & \textbf{P95} & \textbf{Max} \\
\hline
\multicolumn{7}{l}{\emph{Attempts}} \\
4-bit  & 1.00 & 1.07 & 1.37 & 1.61 & 3.28 & 5.88 \\
8-bit  & 1.00 & 1.01 & 1.07 & 1.30 & 2.54 & 4.76 \\
16-bit & 1.00 & 1.01 & 1.09 & 1.33 & 2.04 & 4.76 \\
\hline
\multicolumn{7}{l}{\emph{Time (seconds)}} \\
4-bit  & 1.8 & 1.9 & 2.1 & 2.2 & 3.4 & 5.1 \\
8-bit  & 2.4 & 2.4 & 2.5 & 2.7 & 3.9 & 6.1 \\
16-bit & 4.6 & 4.6 & 4.8 & 5.5 & 7.5 & 15 \\
\hline
\end{tabular}
\end{table}

\begin{figure}[t!]
    \centering
    \includegraphics[width=\linewidth]{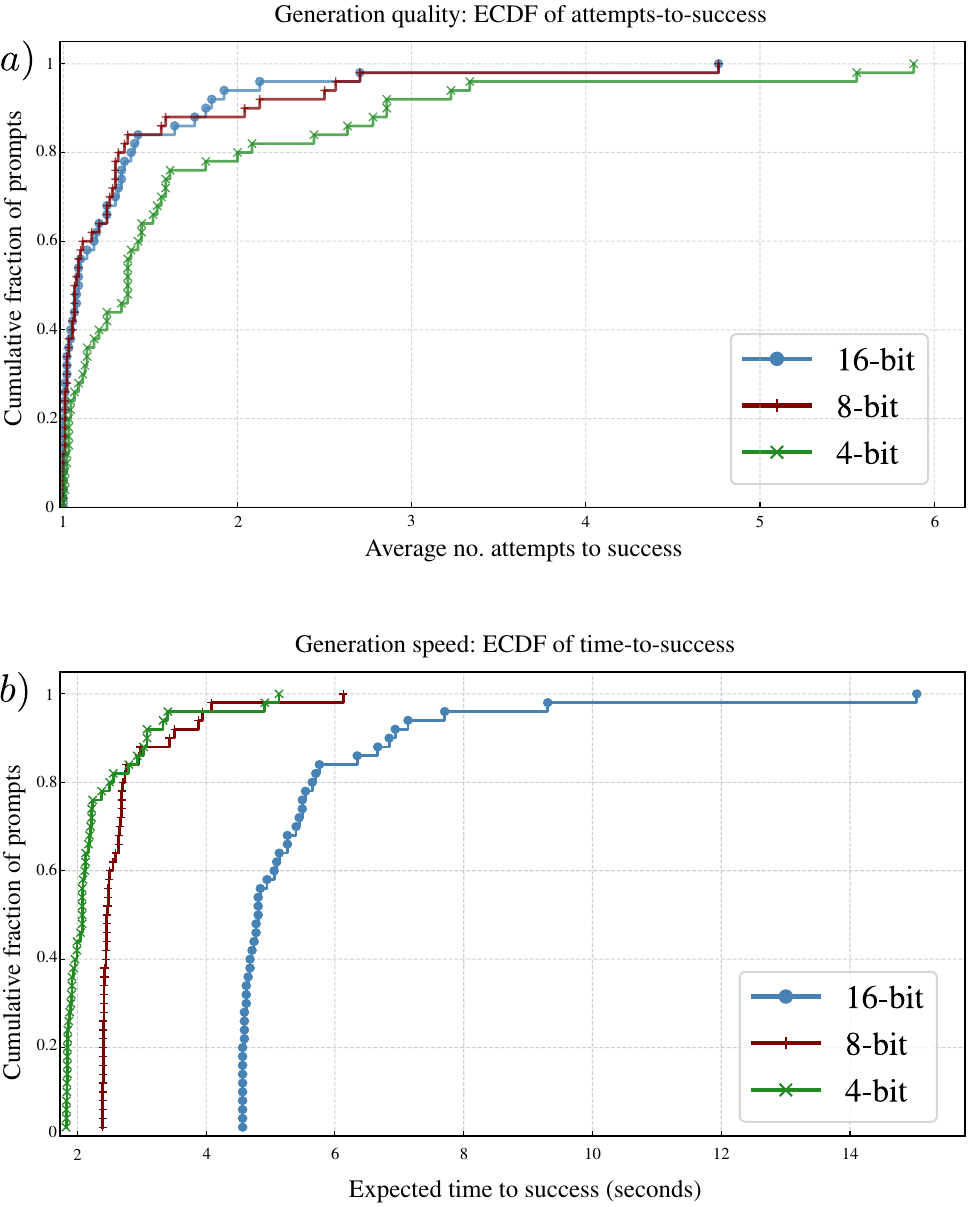}
    \caption{Distribution of generation efficiency across quantization levels.
    \textbf{(a)} ECDF of the average number of generation attempts to success (i.e., the overall judge verdict, defined in Eq.~\eqref{eq:judge_verdict}, is $J(\omega) = 1$) across 50 prompts with 100 samples each ($T = 0.75$). The 16-bit and 8-bit models show nearly identical distributions, with most prompts requiring fewer than 2 attempts on average. The 4-bit model requires more attempts on average across all prompts.
    \textbf{(b)} ECDF of expected time-to-success, defined in Eq.~\ref{eq:TotalTimeDefn}. Despite requiring more attempts, the 4-bit model achieves fastest generation for most prompts due to superior inference speed, with 8-bit close behind. The 16-bit model exhibits the slowest practical performance.}
    \label{fig:ECDFAttemptsAndTiming}
\end{figure}

Figure~\ref{fig:ECDFAttemptsAndTiming}.a shows the empirical cumulative distribution function (ECDF) of expected attempts to success across all 50 prompts. Again, the 16-bit and 8-bit models exhibit nearly identical distributions, with over 80\% of prompts achieving success within 2 attempts on average. The 4-bit model's ECDF curve falls below the other two, indicating a clear drop in performance across all inputs. It requires approximately 2--3 attempts for most prompts, consistent with its reduced estimated overall pass rate calculated above ($78\,\%$ for the 4-bit compared to  approximately $93\,\%$ for the 16-bit).

To compute the expected time to success, we measured generation times on an average gaming PC by current standards (AMD Ryzen~9 7950X CPU, NVIDIA RTX~3070 8GB VRAM) across all 360 test prompts. The models were converted to .gguf format and deployed using a custom SDK built with Llama.cpp~\cite{gerganov2023llamacpp} designed for integration directly into Unreal 5 and Unity. All layers of the model were offloaded to the GPU. The timing consists of two components: preparation time $t_{\text{prep}}$, including model loading and KV cache computation, and per-token evaluation time $t_{\text{eval}}$. Expected time-to-success is then defined as 
\begin{equation}\label{eq:TotalTimeDefn}
    t_{\text{exp},i}^M = \bar{t}_{\text{prep}}^M + W_i^M \cdot \bar{t}_{\text{eval}}^M,
\end{equation}
where $\bar{t}_{\text{prep}}^M$ and $\bar{t}_{\text{eval}}^M$ denote the median preparation and per-token evaluation times for model $M$, respectively. 
For the outputs, the biggest difference in mean output lengths is $6.1 \,\%$ between the 16-bit model and 4-bit model, with 159 and 150 tokens, respectively. The coefficients of variation (ratio of standard deviation to mean) are $7.7\,\%$ and $9.9\,\%$ respectively, so the 4-bit model shows slightly higher variability. While the shorter 4-bit outputs provide a minor speed advantage, this is negligible compared to the dramatic difference in per-token generation speed shown in Table~\ref{tab:timing_breakdown}: $3.6$ ms for the 4-bit model versus $16$ ms for the 16-bit, a speedup of 4.5 times.

Figure~\ref{fig:ECDFAttemptsAndTiming}b reveals how this speedup translates to practical performance. Despite requiring more attempts, the 4-bit model achieves the fastest time-to-success for the vast majority of prompts, with the 8-bit model following closely behind. The percentiles are listed in Table~\ref{tab:time_to_success_percentiles}: the 4-bit and 8-bit models achieve median times of $2.1$\,s and $2.5$\,s respectively, while the 16-bit model reaches $4.8$\,s, with its slowest prompts exceeding $7.5$\,s due to the compounding effect of retries on its slower per-token inference. These results demonstrate that quantization not only maintains acceptable quality but may be necessary for real-time generation scenarios.

All quantization levels show strong rank correlation in overall prompt difficulty. For all pairings, we observed Spearman's $\rho > 0.82$ ($p < 10^{-13}$). Spearman's correlation measures agreement in how models rank-order prompts by difficulty, not absolute score agreement. This indicates that quantized models largely preserve the same ranking of prompt difficulty as their full-precision counterparts. The 8-bit model shows near-perfect agreement with 16-bit ($\rho = 0.93$), while 4-bit quantization shows slightly more deviation ($\rho = 0.83$).
However, a critical question remains: Do these models struggle with the same prompts even among the most difficult inputs, or does quantization introduce different failure modes on the hardest cases? To investigate this, we first pooled success rates across all models and prompts to determine a difficulty threshold, choosing the 40th percentile, which gives a success probability threshold of $80\,\%$. We identified 21 prompts below this threshold. On these, the 16-bit and 8-bit models maintain strong agreement ($\rho = 0.84$, $p < 0.001$), while the 4-bit model shows only weak correlation with both ($\rho = 0.40$ and $0.30$), suggesting 4-bit quantization introduces new failure modes on the hardest prompts.

\section{Discussion and future work}\label{sec:Discussion}
We have demonstrated that a single fine-tuned SLM can service a game loop with a challenging generative task under practical constraints. We now address key limitations that must be resolved for practical deployment, followed by broader implications for future work.

\subsection{Limitations and ongoing Work}\label{sec:Limitations}

\paragraph{Runtime quality assessment}
In this work, we assessed quality fluctuations and their input-dependency using an LLM-as-a-judge framework to establish that adequate quality is achievable with bounded retry requirements. Although this showed the theoretical viability of the SLM in game, actual application at runtime deployment hinges on replacing the cloud-based LLM-as-a-judge with a sufficiently powerful local quality assessment technique, and achieving this is the subject of ongoing research. While we observed no irrecoverable prompts under GPT-4o across 50 test prompts with 100 samples each, stricter quality standards may reveal such cases, in which case the retry-until-success strategy becomes fundamentally unreliable and alternative strategies, such as simplification of the input or batch pre-generation during loading screens, would need to be investigated.

\paragraph{Evaluation methodology}
As noted in Section~\ref{sec:Results}, a concern with using GPT-4o as both teacher and judge is that self-bias may extend irrationally to the SLM derivatives. To assess this, we conducted human annotation on both the outputs of the LLM and all 3 quantization levels of DefameLM for 100 prompts (i.e. $N = 400$). The outputs were scored by the first author on a four-level scale ($0$, $0.33$, $0.66$, $1$) ranging from severely lacking outputs to outputs capturing the task in its totality with no meaningful weaknesses. These scores were used both for per-criterion averages and, after binarization, for paired agreement tests. Note that the human averages are conservative: the four-level scale rarely assigns full marks, widening the apparent gap with binary LLM judges. Furthermore, cross-model validation was performed with Claude Sonnet~4 using multiple prompt variants\footnote{For Claude Sonnet~4, the ``strict judge'' persona instruction was removed and genre context was added (medieval propaganda for an RPG). Applying the same modifications to GPT-4o produced negligible effects ($< 2$\,pp), but for Claude it substantially improved alignment with human annotation.}. To assess the self-bias of GPT-4o, we use the moderate pass criterion for human annotation (i.e. outputs pass when the score is at least $0.66$) and compute the ratio of human-to-judge failure rates. Under GPT-4o this declines monotonically from $27$ (LLM) to $9.8$ (16-bit), $5.1$ (8-bit), and $3.3$ (4-bit). Under Claude Sonnet~4, the same ratio declines from $2.4$ (LLM) to $2.0$ (16-bit), $1.7$ (8-bit), and $1.2$ (4-bit). The steep decline under GPT-4o indicates that its self-bias concentrates on its own outputs rather than extending to the SLM derivatives. Human and Claude Sonnet~4 show the closest agreement (McNemar $\chi^2 = 2.01$, $p = 0.16$; Cohen's $\kappa = 0.284$), with disagreements distributed symmetrically. Human and GPT-4o show weaker agreement ($\kappa = 0.155$) with strongly asymmetric disagreement (McNemar $p < 0.0001$), consistent with GPT-4o's systematic leniency. Since the goal is to measure quality transfer from the teacher LLM, this leniency is acceptable---absolute quality under stricter standards is addressed in the deployment quality discussion below. All capable judges preserve the rank ordering (16-bit $\approx$ 8-bit $>$ 4-bit, LLM highest). The criteria closest to objective task completion (intelligence incorporation, audience targeting, rhetorical targeting) show consistent measurement across judges and human annotation, while alignment, writing quality, and overall assessment show the largest variation between judges. At one extreme, GPT-4o-mini passes nearly all outputs, providing negligible discriminative signal. At the other, Claude Haiku~4.5 collapses on the highly correlated alignment criterion, passing only $22\,\%$ of outputs pooled across all models (human: $95\,\%$), dragging the overall verdict to $20\,\%$. A notable blind spot is writing quality, which passes at $100\,\%$ under GPT-4o but degrades under human annotation and Claude Sonnet~4, particularly for the 4-bit model. Whether this reflects persistent self-bias or simply an insufficiently granular bar on this criterion cannot be determined from the present data. This does not dramatically affect the reported overall verdict but would become a meaningful discriminator under stricter quality standards.

\paragraph{Deployment quality} The objective of the present study was to verify the ability of an SLM to match the quality of its teacher LLM under a retry-until-success strategy. Integration into a specific game will impose a different, potentially higher quality bar depending on the application. To approximate the effect of stricter quality standards, we estimate expected generation times using mean pass rates from human annotation and the timing components from Table~\ref{tab:timing_breakdown} via Eq.~\ref{eq:TotalTimeDefn}. For comparison, under GPT-4o, expected generation times are 2.0\,s (4-bit), 2.5\,s (8-bit), and 4.8\,s (16-bit), close to the measured median times. Under a moderate human threshold (i.e. passed outputs have scores of at least $0.66$, leading to the observed pass rates: 4-bit $53\,\%$, 8-bit $80\,\%$, 16-bit $80\,\%$, LLM $92\,\%$), expected times increase to 2.4\,s (4-bit), 2.6\,s (8-bit), and 5.3\,s (16-bit). Even under a strict human threshold, requiring the maximum score of $1$ on every criterion, pass rates drop substantially (4-bit $20\,\%$, 8-bit $45\,\%$, 16-bit $35\,\%$, LLM $67\,\%$), yet the 4-bit and 8-bit models remain within the 5-second budget at 4.5\,s and 3.6\,s, with the 8-bit model now faster due to its higher pass rate requiring fewer retries, while the 16-bit model exceeds the budget at 9.7\,s. These estimates assume that per-prompt success distributions under stricter standards behave similarly to those under GPT-4o, which remains to be investigated. In particular, the tail may contain irrecoverable prompts. Nevertheless, the 8-bit model emerges as the robust practical choice across all evaluation standards tested.

Manual inspection revealed systematic errors traceable to the training data: certain rhetorical angles were interpreted ambiguously by the teacher LLM, and quantization amplified these patterns in the SLM. Substantial quality improvements are therefore expected from better training data alone, with no changes to the remaining methodology.

\paragraph{Output structure and consistency}
Central to our proposed framework for SLM use in video games is that the models should have a structured input-output format, which serves as a key control for the creativity-consistency trade-off. DefameLM employs a highly structured format, first addressing the target using the chosen intel and style, then appealing to the audience, and finally contrasting with the sender. Initial attempts with less structured outputs resulted in excessively long generations of over 300 words. Imposing the current structure reduced the length to 100--200 words while maintaining quality. A possible concern is that this structure produces stylistic repetitiveness across outputs, which could affect sustained player immersion. The DAG-based data generation approach provides a natural avenue for addressing this by introducing structural variation in the training data, though this was not optimized in the present study, and the degree to which the creativity-consistency trade-off is  surmountable for SLMs remains to be investigated.

\subsection{Broader Implications and Future Directions}
\paragraph{Agentic networks of SLMs}
Just as task decomposition appears necessary for narratively complex LLM-generated content, open-context generation with SLMs will likely require multiple coordinated models. However, the narrow specialization that ensures SLM consistency may limit the ability of complex narratives to emerge, and the degree to which this is possible is a crucial topic for future research. Even granting its feasibility, multi-model coordination introduces additional practical considerations: generation times accumulate, quality assessment requirements grow, and masking behind gameplay sequences becomes more challenging. Our DAG-based data generation approach provides a natural architectural template for runtime coordination, with choice nodes querying game state and generation nodes invoking specialized SLMs. Alternative strategies such as batch-generating content during loading screens or building pre-generated libraries for conditional deployment may be necessary to stay within practical time constraints. Just as training data was generated bottom-up using hierarchies of interdependent information, final model outputs could be assembled similarly, with parallel models handling different subtasks, some generating content, others retrieving game data or using heuristic graphs as in Legashev \emph{et al}.~\cite{Legashev2025NeuroSymbolicDialogue} to reflect character motivations. This modular approach offers advantages in iterative refinement and artistic control. 

Although we have focused on task decomposition for complex open-context tasks, we note that the practicality of single-SLM deployment could yield applications in conjunction with existing PCG frameworks, where an SLM could provide contextual language generation that is difficult to achieve algorithmically.

\paragraph{Implications of data-generation strategy}
The DAG-based data generation approach controls the degree of task specialization. We observed that DefameLM's outputs degrade when inputs contain names or terms not well-represented in the training data, suggesting that insufficient variation at certain choice nodes can cause the model to overfit to specific surface forms rather than learning the underlying patterns. Conversely, when the game world contains a finite set of entities at a given choice node (such as factions), the model can learn facts about individual elements --- a form of world-grounding whose extent and reliability is an interesting topic for future research. While we conjecture that task complexity and required specialization are simply related (higher complexity demands narrower scoping) the DAG approach offers a natural methodology for future empirical investigation of this relationship.

\paragraph{Human oversight and ethical considerations}
The broader gaming community has expressed increasing skepticism towards the use of generative AI~\cite{GamerConcernAIQuanticFoundry2025}, especially when used for creative tasks such as artwork, music or narrative content. We share certain concerns about artistic authenticity and labor displacement, but even though the path we are suggesting here is technical, the most technical aspects, such as hyperparameter selection, can in principle be automated. In this case, successful implementation requires writers and game developers, not data scientists. Indeed,  human creative direction is needed throughout, examples including curating training data, defining quality standards, designing which gameplay events surface as narratively relevant, steering overall aesthetic goals. The creative role shifts from authoring individual pieces to designing the systems and constraints that generate them. Furthermore, for agentic SLM systems targeting open-context generation, the decomposition of narrative tasks is itself a highly non-trivial creative problem and a direct way to impose narratological structure. Similarly, validating that this approach creates engaging player experiences requires human evaluation in complete game systems, which is beyond our current scope but represents necessary future work.
Guard-railing against inappropriate content (slurs, harmful stereotypes) has been achieved at all quantization levels through data curation and training, with no inappropriate outputs observed during human annotation. In particular, we find that the choice of words, phrasings and ideas in the outputs of DefameLM so heavily mimic those of the training data that the technique of aggressive LoRA-fine tuning provides a substantial automatic measure of safeguarding simply through training data curation.
Finally, fine-tuned SLMs offer a path towards transparency in artist rights and fair compensation, since every component of the pipeline (base model, training data, and generation process) can in principle be audited. Applying the framework using only open-data models such as Apertus~\cite{Hernandez2025Apertus} would represent a concrete step in this direction.

\section{Conclusion}\label{sec:Conclusion}
In this study we have presented aggressively fine-tuned SLMs as an alternative to LLMs for dynamic game content generation. As a proof of concept, we introduced DefameLM, a fine-tuned SLM that generates rhetorical attacks between the player and NPCs, requiring synthesis of intelligence items, implementation of a rhetorical angle, and audience-appropriate humor within a constrained format. We benchmarked the model at three quantization levels (16-, 8-, and 4-bit) using a rubric-based LLM-as-a-judge evaluation, measuring quality transfer from the teacher LLM (GPT-4o) to the SLM. The 16- and 8-bit models performed identically ($p = 0.41$), achieving success rates around $93\,\%$, while the 4-bit model performed significantly worse at $78\,\%$. Per-prompt success rates varied substantially across prompts for all models, ranging from approximately $0.2$ to $1$. Our experiments suggest that the 4-bit model develops distinct failure modes on the hardest prompts, beyond the difficulties shared by all three models.

By benchmarking on consumer hardware, we demonstrated that a retry-until-success strategy keeps expected generation times within a 5-second budget, with the 8-bit and 4-bit models close in performance and significantly faster than the 16-bit model. Human annotation and cross-model validation confirm that this holds even under substantially stricter quality standards, with generation time estimates under strict human annotation yielding 3.5\,s (8-bit) and 4.5\,s (4-bit), while the 16-bit model exceeds the budget at 9.7\,s, though per-prompt distributions under stricter standards remain to be verified. This establishes the infrastructure-level feasibility of the approach under practical deployment constraints, with the 8-bit model emerging as the robust practical choice. The critical remaining barrier is local runtime quality assessment to replace cloud-based LLM-as-a-judge evaluation, the demonstration of which would achieve the final verification of the practicality of using SLMs for game content generation.

Our vision is to extend this approach to agentic systems of coordinated, tool-calling SLMs in order to make progress on the harder problem of open-context generation. We believe such networks can provide a scalable foundation for dynamic game content generation across diverse narrative contexts.

\begin{acks}
This work was supported by the Innovation Fund Denmark, grant number 4298-00007B.

The authors thank Mike Iorizzo for developing the SDK used for model deployment and for conducting the benchmarking experiments that measured DefameLM's generation times on consumer hardware.
\end{acks}

\onecolumn

\appendix
 \section{Complete text sample}\label{app:FullTextSample}
Here is the complete text sample shown in redacted form in Fig.~\ref{fig:InputOutputDefameLM}.

\subsection{Input Data}

\par\noindent
\setlength{\fboxsep}{6pt}\setlength{\fboxrule}{0.4pt}%
\colorbox{titlebar}{%
  \makebox[\dimexpr\linewidth-2\fboxsep\relax][l]{\color{white}\textbf{ Input Data}}%
}\par\nointerlineskip
\noindent\fcolorbox{boxborder}{black!4}{%
\begin{minipage}{\dimexpr\linewidth-2\fboxsep-2\fboxrule\relax}
\vspace{2pt}\UseVerbatim[fontsize=\small]{inputdataverb}\vspace{2pt}
\end{minipage}%
}\par\vspace{0.8em}

\subsection{Model Outputs}

\par\noindent
\colorbox{titlebar}{%
  \makebox[\dimexpr\linewidth-2\fboxsep\relax][l]{\color{white}\textbf{ LLM \emph{gold standard} (GPT-4o)}}%
}\par\nointerlineskip
\noindent\fcolorbox{boxborder}{golden}{%
\begin{minipage}{\dimexpr\linewidth-2\fboxsep-2\fboxrule\relax}
\vspace{2pt}\textbf{Freyd\'is the Foolhardy} \\
\emph{The Noblewoman with a Raven's Nest for a Brain}

\vspace{0.5em}
Freyd\'is the Foolhardy, descendant of a long line of blundering nobles, once tried her hand at diplomacy only to turn it into a tavern brawl of words. Her addiction to high-stakes folly has left her with more enemies than allies. Soldiers, do you trust your fate to one who can't even secure an alliance without a tantrum? While she plays her games, your lives hang in the balance. Choose wisely whom you follow.

\vspace{0.5em}
John Kantakouzenos stands for honor and stability, unlike Freyd\'is's reckless lineage.

\vspace{0.75em}
{\large\itshape John Kantakouzenos} \\
\hfill\emph{``For honor and the Golden Chalice!''}
\vspace{2pt}
\end{minipage}%
}\par\vspace{0.8em}

\noindent\begin{minipage}{\linewidth}
\par\noindent
\colorbox{titlebar}{%
  \makebox[\dimexpr\linewidth-2\fboxsep\relax][l]{\color{white}\textbf{ DefameLM: 16-bit}}%
}\par\nointerlineskip
\noindent\fcolorbox{boxborder}{fine16}{%
\begin{minipage}{\dimexpr\linewidth-2\fboxsep-2\fboxrule\relax}
\vspace{2pt}\textbf{Freyd\'is the Foolhardy} \\
\emph{A noblewoman with a lineage of blunders.}

\vspace{0.5em}
Behold Freyd\'is the Foolhardy, whose family tree is as crooked as her failed diplomatic missions! Her ancestors were known more for their blunders than their brilliance, and it seems the Raven's Council has inherited their folly. Soldiers, do you wish to follow a leader who stumbles through politics like a blindfolded jester? While Freyd\'is chases the thrill of high-stakes gambles, our own John Kantakouzenos stands as a paragon of honor and wisdom, guiding us with steadfast courage.

\vspace{0.5em}
John Kantakouzenos: A knight of true lineage, not a lineage of blunders!

\vspace{0.75em}
{\large\itshape John Kantakouzenos} \\
\hfill\emph{``For honor and the Golden Chalice!''}
\vspace{2pt}
\end{minipage}%
}
\end{minipage}\par\vspace{0.8em}\par\vspace{0.8em}

\par\noindent
\colorbox{titlebar}{%
  \makebox[\dimexpr\linewidth-2\fboxsep\relax][l]{\color{white}\textbf{ DefameLM: 8-bit}}%
}\par\nointerlineskip
\noindent\fcolorbox{boxborder}{quant8}{%
\begin{minipage}{\dimexpr\linewidth-2\fboxsep-2\fboxrule\relax}
\vspace{2pt}\textbf{Freyd\'is the Foolhardy} \\
\emph{A noblewoman with a lineage of blunders.}

\vspace{0.5em}
Behold Freyd\'is the Foolhardy, whose family tree is as crooked as her failed diplomatic missions! Her ancestors were known more for their blunders than their brilliance, and it seems the Raven's Council has inherited their folly. Soldiers, do you wish to follow a leader who stumbles through politics like a blindfolded jester? While she chases the thrill of turmoil, you stand guard against chaos. Stand with John Kantakouzenos, whose honor and loyalty are as steadfast as the Golden Chalice.

\vspace{0.5em}
John Kantakouzenos: A knight of true lineage, not a blundering noble!

\vspace{0.75em}
{\large\itshape John Kantakouzenos} \\
\hfill\emph{``For honor and the Golden Chalice!''}
\vspace{2pt}
\end{minipage}%
}\par\vspace{0.8em}

\par\noindent
\colorbox{titlebar}{%
  \makebox[\dimexpr\linewidth-2\fboxsep\relax][l]{\color{white}\textbf{ DefameLM: 4-bit}}%
}\par\nointerlineskip
\noindent\fcolorbox{boxborder}{quant4}{%
\begin{minipage}{\dimexpr\linewidth-2\fboxsep-2\fboxrule\relax}
\vspace{2pt}\textbf{Freyd\'is the Fumbling} \\
\emph{The Noblewoman Who Can't Navigate Her Own Kin}

\vspace{0.5em}
Behold Freyd\'is the Fumbling, whose family tree is as twisted as her political maneuvers! Her noble blood has brewed with folly, leading to a diplomatic disaster and an addiction to the thrill of chaos. Soldiers, do you trust your fate to one who can't even steer her own kin straight? Stand with John Kantakouzenos, whose loyalty to the Order of the Golden Chalice is as steadfast as your courage on the battlefield.

\vspace{0.5em}
John Kantakouzenos: A knight of true lineage, unlike Freyd\'is's tangled roots.

\vspace{0.75em}
{\large\itshape John Kantakouzenos} \\
\hfill\emph{``For honor and the Golden Chalice!''}
\vspace{2pt}
\end{minipage}%
}\par\vspace{0.8em}

\bibliography{bibliography}

\end{document}